\documentclass{article} % For LaTeX2e
\usepackage{iclr2027_conference,times}

\usepackage{amsmath,amsfonts,bm}
\usepackage{fontawesome5}

\def\eqref#1{equation~\ref{#1}}
\def\1{\bm{1}}

\DeclareMathAlphabet{\mathsfit}{\encodingdefault}{\sfdefault}{m}{sl}
\SetMathAlphabet{\mathsfit}{bold}{\encodingdefault}{\sfdefault}{bx}{n}

\usepackage{hyperref}
\usepackage{url}
\usepackage[table]{xcolor}
\usepackage{graphicx}
\usepackage{mathtools}
\usepackage{amssymb}
\usepackage{bbm}
\usepackage{amsmath}
\usepackage{amsthm}
\usepackage{tabularx}
\usepackage{multirow, threeparttable, booktabs, makecell, adjustbox}
\usepackage{caption}
\usepackage{tcolorbox}
\usepackage{graphicx}
\usepackage{subcaption}

\newtheorem{proposition}{Proposition}[section]

\newcommand\bs\boldsymbol
\tcbuselibrary{theorems}
\usepackage{enumitem}
\usepackage[T1]{fontenc}
\usepackage{microtype}
\usepackage{inconsolata}
\usepackage{graphicx}
\usepackage{enumitem}
\usepackage{wrapfig}
\usepackage{makecell}
\usepackage{tabularx}

\usepackage{listings}

\title{Agentic World Modeling for Agent Test-time Scaling without Environment Replica
}
\iclrfinalcopy
\author{%
\begin{tabular}[t]{@{}p{\dimexpr\textwidth-2\tabcolsep\relax}@{}}
\raggedright\large\bfseries
Wendi~Li$^{1,2}$\thanks{Work done during an internship at Microsoft.},
Huseyin~A.~Inan$^{2}$\thanks{These authors contributed equally and share second authorship, listed alphabetically by first name.},
Wei-Ning~Chen$^{2}$\footnotemark[\value{footnote}],
Yanzhi~Chen$^{2}$\footnotemark[\value{footnote}],
Chetan~Bansal$^{2}$,
Saravan~Rajmohan$^{2}$,
Sharon~Li$^{1}$
\tabularnewline[0.5cm]
\normalfont\normalsize
$^{1}$University of Wisconsin--Madison
\qquad
$^{2}$Microsoft
\end{tabular}%
}

\begin{document}

\maketitle

\begin{abstract}

Test-time scaling (TTS) improves large language model agents by sampling and selecting among multiple candidate trajectories, but common realizations assume
that these trajectories can be executed in parallel. This assumption breaks in realistic deployments, where enterprise workspaces and cloud services are shared and persistent, costly or infeasible to replicate, and actions may introduce irreversible side effects or cross-trajectory interference.
Existing generative world models avoid direct execution, but typically compress environment state into a single prompt, limiting their ability to reliably simulate transitions under partial observability.
Instead, we introduce \emph{Agentic World Modeling} (AWM), which reformulates next-observation prediction as a multi-turn evidence-acquisition task.
For each hypothetical action, an agentic simulator adaptively queries the real environment through safe read-only interfaces, reasons over the returned evidence, and uses executable computation to construct the resulting observation without mutating shared state.
Integrated with trajectory-level TTS, AWM enables parallel exploration over a single shared environment.
Across $\tau$-bench, AppWorld, and OfficeBench, AWM outperforms a single-shot simulation baseline by {40.07\%} on average, and surpasses a trained world-model baseline by {12.36\%} on average.
These results establish interactive, read-grounded simulation as a practical foundation for safe and scalable agentic test-time scaling in shared, non-resettable environments. Code:~\href{https://github.com/WindyLee0822/Agentic_World_Model}{\faGithub}
% \end{abstract}
\end{abstract}

\section{Introduction}

Test-time scaling (TTS) \citep{tts,s1,sc} improves LLM agents \citep{sweagent,agent-class1,agent-class2} by sampling multiple candidate trajectories and selecting one for execution. For static reasoning tasks, these trajectories can be sampled in parallel without external-environment interactions. Agentic TTS \citep{gui-tts,agenttti,agenttts,aggregate,aggregate1}, however, requires knowing the observation induced by each hypothetical action to continue its trajectory. Existing approaches obtain these observations by forking or resetting the environment, allowing candidate trajectories to interact with isolated instances.

\begin{figure}
    \centering
    \vspace{-1.2cm}
    \includegraphics[width=0.96\linewidth]{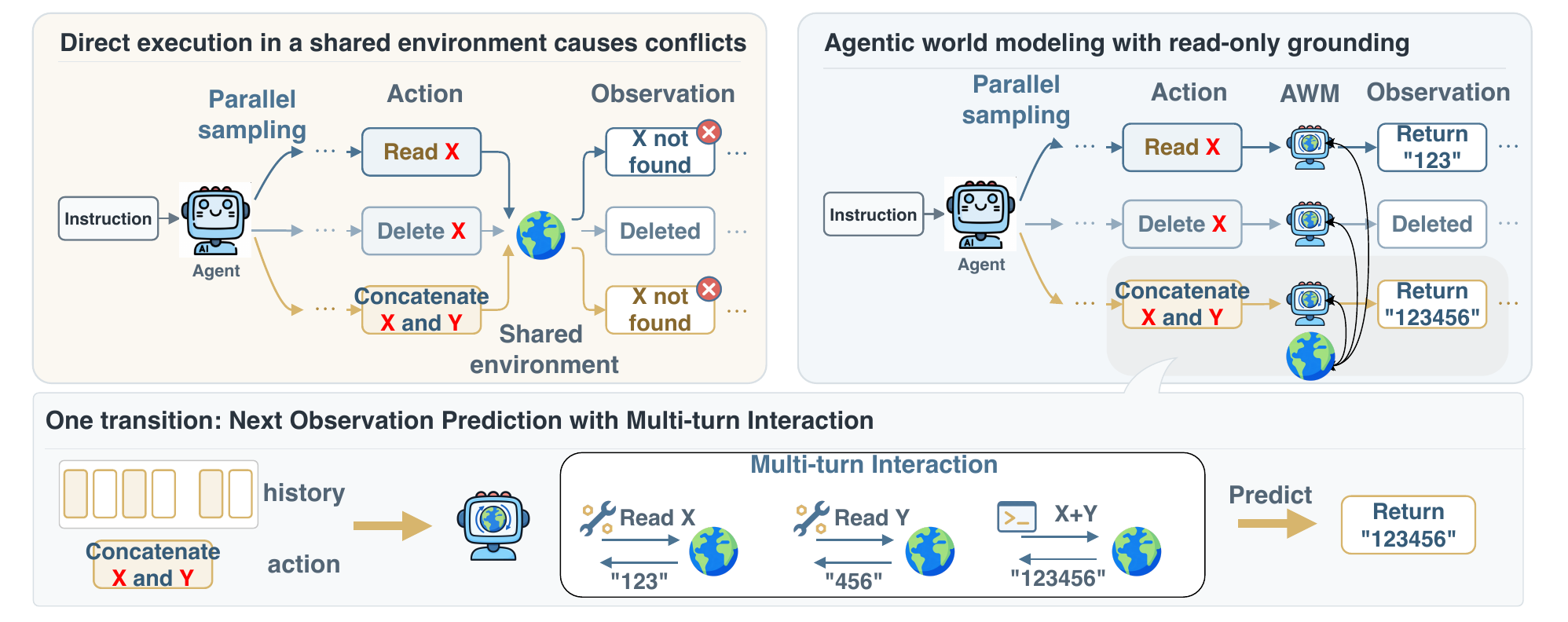}
    \vspace{-0.2cm}
    \caption{{Agentic world modeling (AWM) for test-time scaling in non-replicable environments.}
\textbf{Top left:} Directly executing parallel candidate trajectories in a single live environment causes cross-trajectory interference.
\textbf{Top right:} AWM simulates the environment output with read-only access of real environment to ground prediction.
\textbf{Bottom:} For each transition, the simulator adaptively gathers evidence through multi-turn read-only queries, using computational tools when needed to predict the next observation. 
}
\vspace{-0.4cm}
\label{fig:overview}
\end{figure}

However, such forked replication is often unavailable in realistic systems. For example, an enterprise environment may contain millions of data entries subject to heterogeneous privacy and access restrictions, making faithful replication operationally prohibitive or disallowed altogether. At the same time, multiple candidate trajectories cannot safely share the single live environment through normal execution. Their actions may contain persistent or irreversible mutations \citep{sys,sys1}, such as modifying shared records, changing configurations, or deleting resources. Executing one speculative branch would therefore alter the state seen by the others, creating cross-trajectory interference. As illustrated in the upper-left panel of
Figure~\ref{fig:overview}, executing \texttt{Delete X} in one branch causes
parallel branches that read or use \(X\) to observe that \(X\) no longer
exists, rather than the outcome each branch would have obtained from the
common initial state.

This leaves a fundamental gap: agentic TTS requires the branch-specific observation induced by each candidate action, yet obtaining such observations through real execution on a non-replicable environment may be unsafe. To circumvent this, \emph{world model} \citep{wma,llmwm,agentwm} allows simulating the environment's response to a candidate action without actually executing it or committing the corresponding state mutations. Existing world models \citep{llmwm,llmwm1,simia}, however, often struggle to predict next observations faithfully. To ground their predictions, they typically construct a fixed input context through heuristic retrieval or summarization of the environment state. Yet relevant evidence may be sparse and distributed across a large environment, making it difficult to identify and compress into a single prompt under practical context limits \citep{workspace}. Consequently, the resulting context may omit critical evidence, forcing the world model to produce unsupported or incorrect predictions.

To overcome this limitation, we introduce \emph{agentic world modeling} (AWM), which reformulates next-observation prediction from a single-turn generation problem into a multi-turn, interactive evidence-acquisition agentic process. Rather than reasoning from a fixed, lossy summary of the environment, the simulator identifies missing information, retrieves it through safe read-only interfaces to the real environment, and invokes computational tools when necessary to integrate the retrieved evidence and construct the resulting observation, as shown in the bottom panel of Figure~\ref{fig:overview}. 
We further show theoretically that adaptivity can make evidence acquisition \emph{exponentially} more efficient. This advantage is particularly important in large, interdependent environments where what to inspect next becomes clear only after earlier evidence is observed.

Empirically, we evaluate agentic world modeling on \(\tau\)-bench, AppWorld, and OfficeBench. Across different task complexities and simulator backbones, our approach produces more accurate transition predictions and enables more effective trajectory selection, leading to higher task success than conventional world-modeling baselines. Across all evaluation settings and Best-of-$N$ budgets, AWM outperforms the prompting-based SIMIA baseline by \textbf{40.07 percentage points} on average and, where available, the trained WM-Training baseline by \textbf{12.36 percentage points}. 

Our contributions are threefold:
(1) We formulate next-observation prediction as a multi-turn agentic task in which the simulator adaptively gathers evidence through safe, read-only environment interactions.
(2) We instantiate this formulation using context-isolated subagents and execution-based observation construction, and establish an exponential adaptivity advantage over non-adaptive context construction.
(3) We demonstrate substantial and consistent improvements over prompting-based and trained world models across three agent benchmarks, multiple simulator backbones.

\section{Preliminaries}

\subsection{Partially Observable Markov Decision Process (POMDP)}

We model the environment as a partially observable Markov decision process (POMDP) $\mathcal{M} =(\mathcal{S},\mathcal{A},\mathcal{O},\mathcal{T},\Omega,\gamma)$, where $s_t\in\mathcal{S}$ is the environment state, $a_t\in\mathcal{A}$ is an action, and $o_t\in\mathcal{O}$ is the observation returned to the agent. At step $t$, the agent receives an observation history $h_t=(o_0,a_0,o_1,\ldots,o_t)$ and selects an action under a policy $a_t\sim\pi_\theta(\cdot\mid h_t)$. The latent state then evolves according to the transition dynamics $s_{t+1}\sim \mathcal{T}(\cdot\mid s_t,a_t)$, and emits an observation according to $o_{t+1} \sim \Omega(\cdot\mid s_{t+1})$. For notational convenience, we marginalize over the latent state and fold the transition and observation models into a single \emph{observation-transition model} that acts directly on histories, $o_{t+1} \sim \mathcal{P}(\cdot\mid s_t,a_t)$.

Consider, for example, an enterprise agent operating over a company's digital workspace.
An action $a_t$ might update a customer record or modify a configuration file, while $o_{t+1}$ is the response subsequently exposed to the agent, such as the updated record, a success message, or an error. 
The latent state $s_t$ is the overall state of the whole workspace, which may comprise the contents of millions of files and documents, service configurations, and system logs. Consequently, explicitly representing $s_t$ in an LLM context, is generally infeasible due to both its scale and access constraints.

\subsection{World Modeling for LLM Agents}
 
World modeling predicts how the environment responds to an agent action. Given the current latent state \(s_t\) and action \(a_t\), the environment first transitions according to \(\mathcal{T}(\cdot \mid s_t,a_t)\) and then emits an observation according to \(\Omega(\cdot \mid s_{t+1})\). Their composition defines the true next-observation distribution:
\begin{equation}
    \mathcal{P}(o_{t+1}\mid s_t,a_t)
    =
    \int_{\mathcal{S}}
    \Omega(o_{t+1}\mid s_{t+1})
    \mathcal{T}(d s_{t+1}\mid s_t,a_t).
\end{equation}
A world model \(\mathcal{P}_\theta\) seeks to approximate this distribution without executing \(a_t\) in the real environment.
Prior work on world modeling for LLM agents typically formulates this prediction as a single-turn generation problem. Because the latent state \(s_t\) is not directly available to the language model, it is represented by a statically constructed context, such as a verbalized state description or a set of heuristically retrieved artifacts. The next observation is then generated directly as
\begin{equation}
    o_{t+1}
    \sim
    \mathcal{P}_\theta(\cdot\mid s_t,a_t)
    :=
    \pi_\theta\!\left(\cdot\mid c(s_t),a_t\right).
\end{equation}
In realistic environments, however, \(s_t\) may be too large to serialize faithfully into a single prompt. Consequently, the fixed context may omit transition-relevant evidence while including substantial irrelevant information, leaving the simulator with an incomplete and poorly targeted basis for predicting \(o_{t+1}\).

\section{Agentic World Modeling: Next-State Prediction as a Multi-turn Interactive Task}

To overcome the limitations of fixed, lossy context construction, we make the prediction process itself interactive. Specifically, we reformulate next-observation prediction as an \emph{interactive evidence-acquisition process}: given $(h_t,a_t)$, the simulator adaptively queries safe read-only interfaces, reasons over the retrieved evidence, and produces \(o_{t+1}\) once sufficient information has been gathered. We first formalize this process as a multi-step auxiliary POMDP, then present its practical instantiation, and finally establish its theoretical advantage over conventional single-shot prediction.

\subsection{Next-observation Prediction as a Multi-turn Agentic Task}

In this subsection, we formalize next-observation prediction as a multi-turn agentic task. We first motivate the transition from single-shot generation to interactive evidence acquisition, and then formulate the resulting test-time prediction process as an auxiliary POMDP. This formulation captures how, within a single inference episode, the simulator adaptively gathers evidence through safe read-only interactions, determines when sufficient information has been acquired, and terminates with a predicted next observation.

\paragraph{From single-shot generation to evidence-seeking.}
Rather than requiring the simulator to infer an environment transition solely from its parametric knowledge and a partial or summarized description of the current state, we allow it to actively query the true environment for relevant evidence before committing to a prediction. We utilize the fact that read-only interaction is both safe and naturally parallelizable: inspecting the environment does not modify its state, so evidence for multiple hypothetical trajectories can be collected concurrently without introducing side effects or inconsistencies between rollouts.
Therefore, we equip the simulator with a set of safe read-only tools $\mathcal{R}$, such as file inspection, metadata retrieval, database lookup, and log queries. For each candidate trajectory, the simulator can adaptively invoke these tools, interleave evidence retrieval with reasoning, and terminate the process only when it judges the accumulated evidence sufficient to predict the next observation. Because different candidate rollouts interact with the environment exclusively through non-mutating reads, their evidence-acquisition processes can proceed in parallel while remaining grounded in the same underlying environment state. Consequently, next-observation prediction is no longer based solely on parametric knowledge or incomplete context, but instead becomes an evidence-grounded inference process that is both safely parallelizable and less prone to hallucinated or internally inconsistent transitions.

\paragraph{Formulation of next-observation prediction.}
We formulate the prediction of each environment transition, $o_{t+1}\sim\mathcal{P}(\cdot\mid s_t,a_t)$, as an auxiliary POMDP
$\widetilde{\mathcal{M}}=(\widetilde{\mathcal{S}},
\widetilde{\mathcal{A}},
\widetilde{\mathcal{O}},
\widetilde{\mathcal{T}},
\widetilde{\Omega}).$
Unlike the original environment POMDP, whose latent state evolves as the agent executes actions in the environment, speculative actions in a simulated trajectory are never applied to the real environment. 
At simulator step $i$, the auxiliary state is therefore represented as $ \tilde{s}_i=(s_0,h_t,a_t,\xi_i), $ where $\xi_i=(\tilde a_1,\tilde o_1,\ldots,\tilde a_{i-1},\tilde o_{i-1})$ denotes the evidence-acquisition history accumulated so far. Note that \(t\) indexes transitions in the original agent POMDP, while \(i\) indexes evidence-acquisition steps within the simulator POMDP throughout.
The base environment state $s_0$ remains fixed throughout the auxiliary episode because all intermediate interactions are read-only; only the evidence history $\xi_i$ evolves. Note that the simulator does not materialize or directly access $s_t$; 
it infers the consequences of preceding hypothetical actions from $h_t$ while using read-only access to $s_0$ to acquire factual evidence about the underlying environment.
At each step, the simulator selects an auxiliary action
$
\tilde a_i\sim
\pi_\theta(\cdot\mid \tilde s_i),\ \tilde a_i\in\widetilde{\mathcal A},
$
where $\widetilde{\mathcal A}$ contains both read-only queries and terminal
next-observation predictions. When $\tilde a_i$ is a read action, the auxiliary
state transition appends the query to the interaction history while leaving
$s_0$ unchanged, and the observation model $\widetilde{\Omega}$ returns the
corresponding evidence $\tilde o_i\in\widetilde{\mathcal O}$.
In our setting, each intermediate $\tilde a_i$ invokes a safe read-only tool and $\tilde o_i$ is its returned result, thereby extending the evidence history to $\xi_{i+1}=(\xi_i,\tilde a_i,\tilde o_i)$.
We impose a maximum budget of $M$ read actions. The simulator therefore adaptively gathers an evidence trajectory $\xi_I$, with $0\leq I\leq M$, before terminating. At any step before the budget is exhausted, it may stop evidence acquisition by selecting a terminal prediction action; after $M$ reads, prediction becomes mandatory. The terminal action commits to the predicted next observation of the environment transition, $ o_{t+1} \sim \pi_\theta(\cdot\mid h_t,a_t,\xi_I),$ and ends the auxiliary episode.
Thus, next-observation prediction itself becomes a sequential decision-making problem in which the simulator jointly determines \emph{what evidence to acquire} and \emph{when sufficient evidence has been gathered} to make the final prediction.

\subsection{Practical Instantiation within Test-Time Scaling}

Instantiating agentic world modeling within test-time scaling introduces two
practical challenges. First, each simulated trajectory nests multiple
evidence-gathering episodes, so naively retaining every intermediate interaction
would make the context grow with both rollout length and reasoning depth.
Second, although read-only access can expose the correct environment state, it
does not guarantee that long or structured observations will be reproduced
faithfully. We address these challenges through hierarchical subagent
delegation and a read--compute interface that combines safe environment access
with executable prediction.

\paragraph{Subagents for context isolation.}
A monolithic world model would accumulate the queries, retrieved evidence, and
intermediate reasoning required for every transition, rapidly overwhelming the
trajectory context. We instead realize the world model as a hierarchical
inference process that can invoke context-isolated subagents for bounded
subtasks, such as predicting the observation of a proposed action or verifying
the success of a completed trajectory. Each subagent performs its own
multi-turn evidence gathering and returns only a compact task-specific
result---for example, a predicted observation or a verification score---to the
outer process. The simulator retains the causal consequences needed for later
predictions, but not the full internal interaction trace. This design
decouples computational depth from context length: the world model can perform
substantial internal reasoning while exposing only the information necessary
to continue the simulated trajectory.

\paragraph{Read-only grounding and executable prediction.}
To ground these subagents in the actual environment, we provide a
backend-specific read-only capability set $\mathcal{R}_e \subseteq\{ \textsc{Read},\textsc{List},\textsc{Call},\textsc{Trace}\}.$
Here, \textsc{Read} retrieves the exact content of a specified item, \textsc{List} enumerates relevant entities or resources, \textsc{Call} executes a whitelisted non-mutating environment query, and \textsc{Trace} reveals the provenance or simulated dependencies associated with a resource. Together, these tools let the world model acquire transition-relevant evidence
without modifying the underlying environment.
Read access alone, however, does not ensure faithful output construction. When the next observation is a long string or structured object, asking the model to regenerate retrieved content may introduce omissions, formatting changes, or hallucinated values. We therefore allow restricted Python computation over the retrieved evidence. For deterministic transformations, such as replacing substrings in a document, the subagent can perform the
transformation programmatically and use the program output directly as the
predicted observation. Python thus acts as a decoder: the model determines the required operation, while execution produces the exact result. By bypassing unnecessary free-form regeneration, this mechanism reduces hallucination while preserving the content and structure of the underlying evidence.

\subsection{Theoretical Advantage of Adaptive Evidence Acquisition}
\label{sec:theory}

The key distinction between AWM and conventional context construction is \emph{adaptivity}. Consider a general non-adaptive baseline that, given $(h_t,a_t)$, selects at most $M$ pieces of environment information before observing any of their contents, and then arbitrarily filters, summarizes, or combines the retrieved information to predict $o_{t+1}$. 
AWM instead acquires evidence sequentially. At simulator step $i$, it selects
$\tilde a_i\sim \pi_\theta(\cdot\mid h_t,a_t,\xi_i),$ where $\xi_i=(\tilde a_1,\tilde o_1,\ldots, \tilde a_{i-1},\tilde o_{i-1}),$ so each returned observation $\tilde o_i$ can determine what information should be queried next. 

The advantage of adaptivity becomes pronounced when relevant information must be located through successive dependencies.  
For example, an initial customer record may reveal which account to inspect; that account identifies the applicable policy; and the policy points to the exception that ultimately determines the next observation. Before reading one item, the simulator hardly knows which item will be relevant next. Instead, AWM can naturally follow this chain one step at a time.
% while a non-adaptive
method must anticipate the entire chain in advance. We validate this advantage with the following formalization.

\begin{proposition}[Exponential adaptivity gap, see Appendix~\ref{app:theory} for detailed proofs.]
\label{prop:adaptive-gap}
Consider a transition whose relevant information
$W^\star\in\{0,1\}^{\beta}$ is reached through $D$ successive lookups, where
each retrieved item reveals one relevant next item among $K$ possibilities.
Before observing any evidence, there are therefore $K^D$ possible information
paths.
To achieve prediction error at most $\delta$, the read budgets of adaptive our appraoch $M_{\mathrm{AD}}$ and of non-adaptive baselines $M_{\mathrm{NA}}$ satisfy 
\begin{equation}
M_{\mathrm{AD}}(\delta)\le D+1,
\qquad
M_{\mathrm{NA}}(\delta)
\ge
\left(1-\delta-\frac{1}{\beta}\right)K^D .    
\end{equation}
Thus, the agentic world modeling with adaptive evidence acquisition requires only $O(D)$ reads, while previous paradigms with non-adaptive context construction can require $\Omega(K^D)$ reads for the same prediction accuracy.
\end{proposition}

The result highlights why adaptivity becomes increasingly valuable in large,
interdependent environments. A non-adaptive summarizer or retriever must
anticipate which information will matter before observing the evidence and
therefore spread a limited access budget across many possible dependencies.
AWM instead uses the environment itself to progressively resolve this
uncertainty: each $\tilde o_i$ informs the next $\tilde a_{i+1}$, concentrating
computation on the particular information path relevant to the current
transition. Consequently, AWM can extract substantially more
transition-relevant information from the same environment-access budget,
providing a more efficient and generalizable basis for faithful
next-observation prediction.

\section{Experiments}

\definecolor{sealbg}{HTML}{EEF4FF}
\renewcommand{\arraystretch}{1.1} 
\begin{table}[t!]
\small
    \centering
    \caption{Main results of world-model-guided test-time scaling on $\tau$-bench, AppWorld, and OfficeBench under different Best-of-$N$ (BoN) budgets and simulator backbones. AWM denotes our agentic world model; SIMIA~\citep{simia} and WM-Training~\citep{qwenworld} are prompting-based and training-based world-model baselines, respectively.}
    \label{tab:main_results}
    \begin{tabular}{cc|cc|ccc|ccc}
    \toprule
    \multirow{2}{*}{Methods}&\multirow{2}{*}{BoN}  & \multicolumn{2}{c}{$\tau$-bench} & \multicolumn{3}{c}{AppWorld} &\multicolumn{3}{c}{OfficeBench}   \\
    & & Airline& Retail&   Level1 & Level2& Level3& 1App&2Apps&3Apps \\
    \midrule\midrule
    % \rowcolor{sealbg}\multicolumn{10}{c}{\textsc{GPT-4.1}}\\
    % % Greedy& - & \\ 
    % \multirow{3}{*}{\makecell[c]{\small SIMIA\\ \footnotesize \citep{simia}}}& @1 &44.00&5.26& 25.58\\ 
    % &@4&44.00& 5.26& 35.66\\
    % &@16&46.00&5.26& 39.53\\
    % \midrule
    % \multirow{3}{*}{AWM}& @1 &36.00&42.98& 62.79\\ 
    % &@4 & 56.00&54.39& 71.32\\
    % &@16& 68.00&70.18& 73.64\\
    % \midrule\midrule
  \rowcolor{sealbg}\multicolumn{10}{c}{\textsc{Deepseek-V4-Pro}}\\
    % Greedy& - & \\ 
    \multirow{3}{*}{\makecell[c]{SIMIA\\ \footnotesize \citep{simia}}}& @1 &46.00&9.65& 25.58& 13.13&2.33 &12.90&10.53& 6.25\\ 
    &@4&54.00&13.16& 32.56&16.67 & 4.65 &21.51&16.84& 10.71 \\
    &@16&60.00&19.30& 37.98& 23.23& 8.53&27.96& 24.21& 16.07 \\
    \midrule
    \multirow{3}{*}{\textbf{AWM}}& @1 &62.00&66.67&57.36 &55.56 &33.33 &52.69&43.16&12.50\\ 
    &@4&78.00&78.95&75.19 &78.79 & 44.96&66.67&54.74&17.86\\
    &@16&86.00& 89.47& 90.70& 88.89& 63.57&77.42&69.47& 23.21\\
    \midrule\midrule
    
    \rowcolor{sealbg}\multicolumn{10}{c}{\textsc{GPT-5.4}}\\
    % Greedy& - & \\ 
    \multirow{3}{*}{\makecell[c]{SIMIA\\ \footnotesize \citep{simia}}}& @1 &40.00&5.26 &20.93&7.07&1.55&15.05&9.47&1.79\\ 
    &@4& 46.00&6.14&25.58&9.09&3.10&17.20& 12.63&2.68\\
    &@16&50.00&10.53&31.01&11.11&6.20&27.96&18.95&4.46\\
    \midrule
    \multirow{3}{*}{\textbf{AWM}}& @1 &60.00&49.12&82.17&52.02&37.60& 56.99&53.68&25.89\\ 
    &@4&72.00&71.93&85.27&58.59&47.29& 60.22&61.05&32.14\\
    &@16& 82.00&77.19& 87.60&65.66&53.88&63.44&63.16&34.82\\
    \midrule\midrule
    \rowcolor{sealbg}\multicolumn{10}{c}{\textsc{Qwen3.5-35B-A3B}}\\
    % Greedy& - &  \\
     \multirow{3}{*}{\makecell[c]{SIMIA\\ \footnotesize \citep{simia}}}& @1 &36.00& 7.02&17.05 &5.56 & 1.55&11.83&8.42& 2.68\\ 
    &@4&44.00&10.53&23.26 & 10.10&5.43 &17.20 & 17.89& 4.46\\
    &@16& 52.00&14.04&31.78 &12.63 & 6.20&24.73&21.05& 7.14\\
    \midrule
     \multirow{3}{*}{\makecell[c]{WM-Training\\ \footnotesize \citep{qwenworld}}}& @1 &54.00 &36.84& 17.83&17.17 &7.36 &60.22&43.16&14.29\\ 
    &@4&58.00 &52.63 &40.31 &24.24 &9.30 &66.67&60.00& 24.11\\
    &@16& 64.00&62.28 & 53.49& 32.32& 13.18&67.74&65.26& 33.04\\
    \midrule
    \multirow{3}{*}{\textbf{AWM}}& @1 &66.00&63.16&40.31 &28.79 &13.57 &61.29&62.11&21.43\\ 
    &@4&70.00&75.44&53.49 &34.85 &18.99 &69.89&68.42&35.71\\
    &@16&76.00&79.82& 64.34& 48.48& 33.33&72.04&74.74&41.96 \\
    \bottomrule
    \end{tabular}
\end{table}

% \begin{table}[t]
%     \centering
%     \caption{Caption}
%     \label{tab:placeholder}
%     \scalebox{0.75}{
%     \begin{tabular}{c|ccc|ccc|ccc|ccc|ccc}
%     \toprule
%        \multirow{3}{*}{Methods}  & \multicolumn{12}{c|}{AppWorld} & \multicolumn{9}{c}{Tau} \\
%          & \multicolumn{3}{c|}{Retail}& \multicolumn{3}{c|}{Airline}& \multicolumn{3}{c|}{Single App}& \multicolumn{3}{c|}{Two Apps}& \multicolumn{3}{c}{Three Apps} \\
%         & @1& @4 & @16 & @1& @4 & @16 & @1& @4 & @16 & @1& @4 & @16 & @1& @4 & @16 \\
%                  \midrule
%         \multicolumn{16}{c}{GPT-4.1 (Pass@1 = with \textbf{real} environment) } \\

%     LLM-Sim & \\ 
%     Pass@K & 46.0&66.0 &78.0 & 69.30&88.60 &98.25 &59.14& 65.59& 67.74& 65.75& 79.45& 83.56& 22.08& 31.17& 38.96 \\
%     AWM & 42.0&64.0 &72.0 &64.04 &84.32 &83.33 & 58.06& 61.29&65.59& 74.64& 81.69&83.10& 26.42&35.85& 41.51\\
%     AWM+PE &\\ 
%     \midrule

%     \multicolumn{16}{c}{GPT-5.4 (Pass@1 = with \textbf{real} environment) } \\

%     LLM-Sim & \\
%     Pass@k & 65.49 &92.92&97.35 & 63.27& 83.67& 87.76& 59.14& 61.29&65.59& 64.21&75.79& 83.16& 29.46& 35.71& 41.07\\
%     AWM & 64.91 & 84.21 &92.11 & 64.00&74.00  &86.00 & 61.29 & 63.44& 66.67& 70.21& 75.53& 79.79& 25.00  \\
%     AWM+PE &\\
%     \midrule
%     \multicolumn{16}{c}{Qwen3.5-35B-A3B (Pass@1 = with \textbf{real} environment) } \\

%     LLM-Sim & \\
%     SFT\&RL & \\
%     AWM & \\
%     AWM+PE &\\
    
%     \bottomrule 
%     \end{tabular}}

% \end{table}

\subsection{Experimental Settings}
We evaluate agentic world modeling (AWM) on three agentic benchmarks spanning distinct interaction modalities, and compare against both training-based and prompting-based world-model baselines under a unified protocol.

\paragraph{Benchmarks \& baselines.}
We consider three complementary environments. \textbf{$\tau$-bench}~\citep{taubench} evaluates tool–agent–user interaction in two customer-service domains (retail and airline): the agent must converse with an LLM-simulated user, follow domain-specific policy documents, and issue API tool calls. \textbf{AppWorld}~\citep{appworld} provides a high-fidelity simulated ecosystem of nine everyday applications exposed through 457 APIs, with 750 tasks grounded in the digital activities of roughly 100 fictitious users; it requires agents to iteratively generate and execute code, coordinate actions across applications, and complete complex workflows without causing unintended state changes.
\textbf{OfficeBench}~\citep{officebench} evaluates office automation in an integrated environment comprising applications such as a word processor, spreadsheet editor, and email client. Its 300 tasks span one to three applications and require agents to plan over long horizons, switch between applications at appropriate points, and manipulate application-specific content.
We compare the agentic world modeling (AWM) against two representative world-model paradigms. \textbf{SIMIA}~\citep{simia} (in-context simulation) is a purely generative baseline that verbalizes the environment into the prompt and predicts the next observation in a single shot. \textbf{WM-Training}~\citep{qwenworld} learns environment dynamics through a dedicated training process using the three-stage continued-pretraining, SFT, and RL recipe of \citet{qwenworld}.

\paragraph{Implementation details.}
We set the sampling temperature to \(1.0\) wherever supported and use the agent itself to predict observations and assess trajectory correctness. We conduct experiments with three models: \textbf{DeepSeek-V4-Pro}, an open-weight frontier model with strong general-purpose reasoning and tool-use capabilities;; \textbf{GPT-5.4}, OpenAI's proprietary frontier reasoning model with a \(1\)M-token context window and built-in support for tool and computer use; and \textbf{Qwen3.5-35B-A3B}, an open-weight Mixture-of-Experts model (\(35\)B total / \(3\)B active parameters, \(256\)K context). For the WM-Training baseline of the Qwen model, we use the open-sourced \textbf{Qwen-AgentWorld-35B-A3B}, which is built on Qwen3.5-35B-A3B through continual pretraining, fine-tuning, and post-training on world-modeling data. 
For all models, we cap read-only interaction at six steps and report the real-environment task success rate of the trajectory selected by Best-of-\(N\).

\begin{figure}[t!]
    \centering
\includegraphics[width=\linewidth]{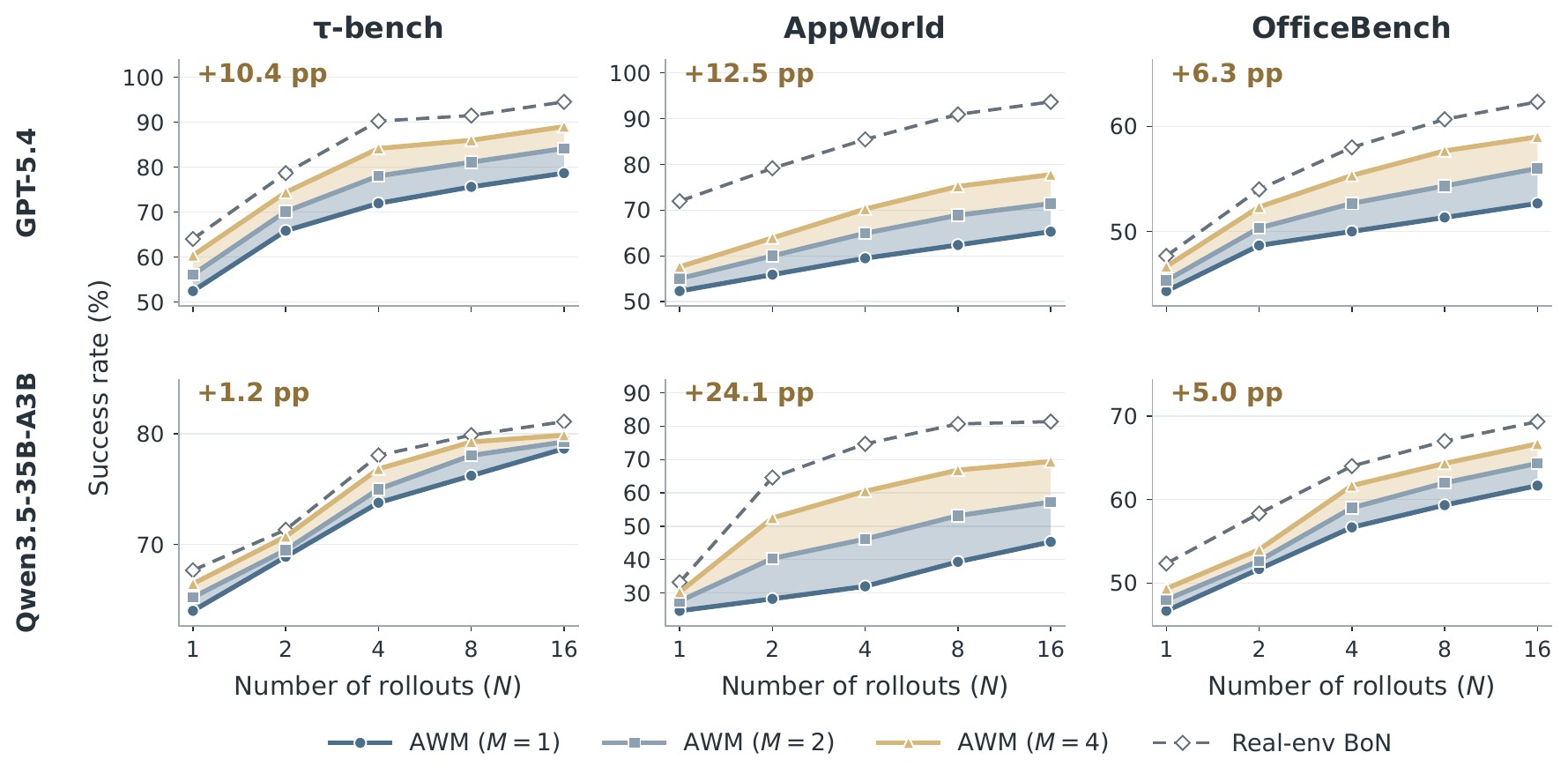}
\vspace{-0.8cm}
    \caption{Scaling simulator-side test-time compute further closes the gap to real-environment performance, consistently improving performance across models and benchmarks. Bold numbers show the performance gain from M=1 to M=4, with N=16 fixed.}
    \label{fig:simulation_scaling}
    \vspace{-0.2cm}
\end{figure}
\begin{figure}[t!]
    \centering
\includegraphics[width=\linewidth]{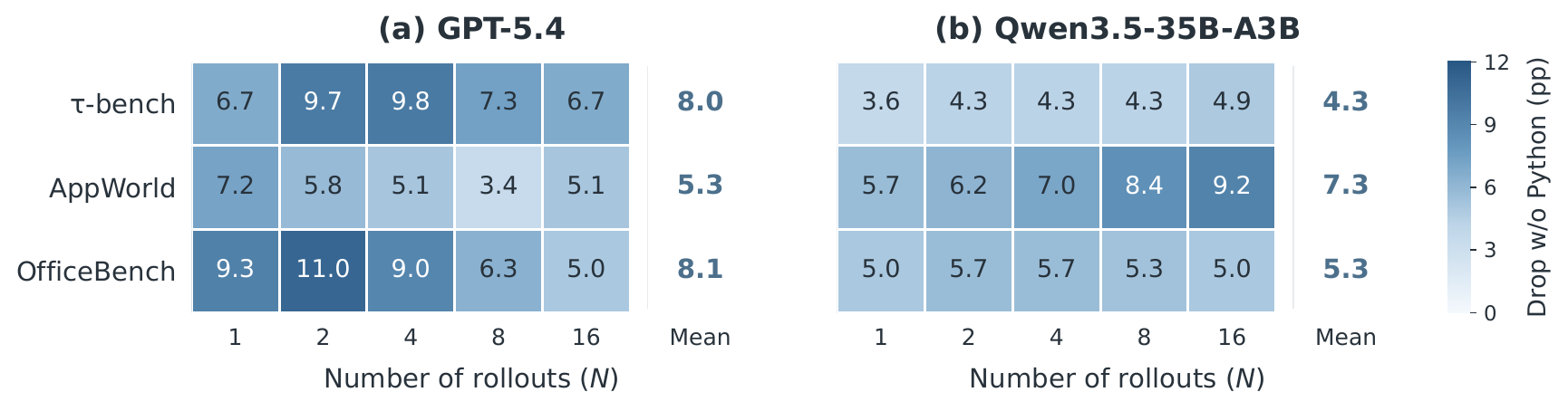}
\vspace{-0.5cm}
    \caption{Ablation of the program tool.
We compare AWM with and without the Python-based program tool across rollout budgets $N$.
Removing the program tool consistently reduces success rates across all three benchmarks and both simulator backbones; the annotations report the average performance drop across $N$.}
\vspace{-0.5cm}
    \label{fig:tool_ablation}
\end{figure}

\subsection{Main Results}

\paragraph{AWM substantially improves trajectory selection.}
AWM consistently outperforms the prompting-based SIMIA baseline across all benchmarks, simulator backbones, and BoN budgets, winning all comparisons in Table~\ref{tab:main_results}.
Macro-averaged over the eight evaluation settings, AWM improves success rate over SIMIA by 40.07 percentage points. The gains are often substantial: with DeepSeek-V4-Pro at BoN@16, AWM improves \(\tau\)-bench Retail from 19.30 to 89.47 (+70.18 points); with GPT-5.4 at BoN@1, it improves AppWorld Level 1 from 20.93 to 82.17 (+61.24 points); and with Qwen3.5-35B-A3B at BoN@16, it improves OfficeBench 2Apps from 21.05 to 74.74 (+53.68 points).
These results show that adaptively acquiring transition-relevant evidence produces simulated trajectories that are substantially more useful for test-time decision making than single-shot world-model prediction.

\paragraph{AWM also outperforms a trained world model.}
For Qwen3.5-35B-A3B, where the training-based WM-Training baseline is available, AWM outperforms WM-Training in all comparisons, by 12.36 percentage points on average.
The improvement is consistent across environments and persists as the BoN budget increases, including gains of +26.32 points on \(\tau\)-bench Retail at BoN@1, +20.16 points on AppWorld Level 3 at BoN@16, and +8.93 points on OfficeBench 3Apps at BoN@16.
This result suggests that adaptive, read-only grounding can provide a stronger basis for environment simulation than a separately trained transition model.

\paragraph{The advantage becomes larger on average with increasing test-time compute.} AWM also makes more effective use of additional trajectory samples.
Averaged across backbones and evaluation settings, its advantage over SIMIA grows from 34.99 points at BoN@1 to 41.04 points at BoN@4 and 44.17 points at BoN@16. Correspondingly, the macro-average success rate of AWM increases from 48.22 at BoN@1 to 66.97 at BoN@16, a gain of 18.74 points, whereas SIMIA increases from 13.23 to 22.79 (+9.56 points).
This widening gap indicates that scaling the number of candidate trajectories is most effective when their consequences can be simulated with sufficient fidelity: additional candidates provide limited benefit when selection is based on inaccurate world-model predictions.

\subsection{Analyses}

\paragraph{Scaling simulator-side test-time compute.}
Beyond increasing the number of candidate rollouts $N$, our framework provides a complementary scaling axis by allocating more computation to each simulated transition. Specifically, instead of relying on a single simulated outcome, we sample $M$ independent predictions and let the simulator aggregate them into the resulting transition, with $M=1$ recovering our default setting.
As shown in Figure~\ref{fig:simulation_scaling}, increasing $M$ from $1$ to $2$ and $4$ consistently improves task success across all three benchmarks and both model backbones. The gains are particularly pronounced in more challenging settings, while the $M=4$ curves systematically move closer to the Real-env BoN ceiling. These results suggest that a meaningful portion of the remaining simulation error can be reduced through additional test-time computation, and that scaling simulator-side inference complements scaling the number of candidate rollouts $N$.

\begin{figure}
    \centering
    \includegraphics[width=\linewidth]{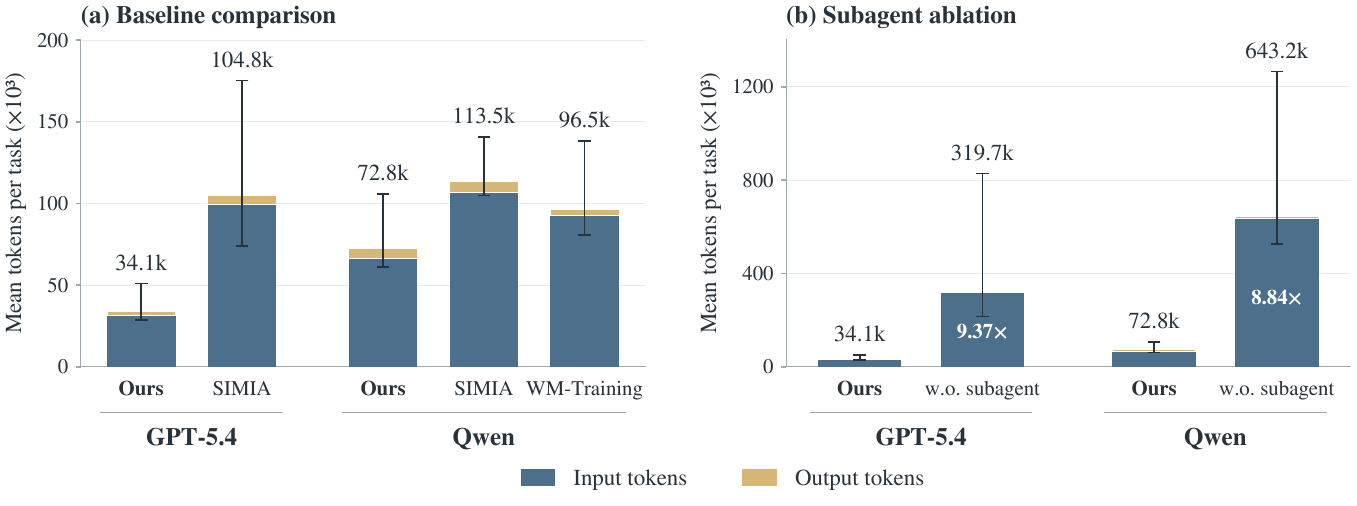}
    \vspace{-0.5cm}
    \caption{
Token-cost comparison and subagent context-management ablation on OfficeBench with GPT-5.4 and Qwen3.5-35B-A3B. Bars report the mean cumulative input and output tokens per task, with whiskers denoting the estimated low--high range.
} \vspace{-0.5cm}
    \label{fig:cost}
\end{figure}

\paragraph{Program tools improve simulation quality.}
The read-only interfaces used by AWM are not intended to prescribe a fixed tool
set: they can be flexibly adapted to the structure and capabilities of each environment. Our implementation therefore serves as one practical prototype, where, in addition to environment-specific read operations, a Python-based program tool for intermediate computation and deterministic processing of retrieved evidence is important.
As shown in Figure~\ref{fig:tool_ablation}, removing this program tool consistently degrades task success across all three benchmarks and both model backbones.
The degradation persists as $N$ increases, indicating that sampling more candidate trajectories does not fully compensate for less reliable transition simulation.
These results suggest that programmatic computation is a beneficial component of our current instantiation, in addition to an evidence-acquisition, read-only interface to an agentic world model.

\paragraph{AWM remains cost-efficient through subagent context management.}
A potential concern with interactive world modeling is that multi-turn evidence acquisition may substantially increase inference cost relative to conventional single-shot simulators. Figure~\ref{fig:cost} shows that this is not the case in our implementation. We extract 64 data from OfficeBench, AWM operates at a token cost comparable to existing world-modeling approaches and is substantially cheaper in the evaluated settings.
This efficiency is enabled in large part by our subagent-based context management, which isolates the evidence and reasoning associated with each simulated transition rather than repeatedly carrying the entire nested interaction history forward. Removing this mechanism increases token usage by $9.37\times$ for GPT-5.4 and $8.84\times$ for Qwen. These results indicate that interactive simulation need not introduce prohibitive overhead: with appropriate context isolation, AWM can retain the benefits of multi-turn evidence acquisition while keeping inference cost in the same regime as, or below, prior methods.

\section{Related Work}

\paragraph{Test-time scaling for agents.}
Test-time scaling (TTS) improves model performance by allocating additional computation at inference time \citep{tts,s1}. Early methods \citep{sc, tot, verifier, prm, pqm, implicitprm} focused primarily on self-contained reasoning tasks with sampling, voting, and verifiers. Recent work extends these ideas to interactive agents by generating, searching, and selecting among candidate action trajectories \citep{treeagent,treeagent1,gui-tts,agenttts,webagent-tts,codeagent-tts,aggregate,aggregate1}. Other approaches allocate computation through reflection, critique, verification, and sequential revision \citep{reflexion, critic,wan2026inference}. Together, these studies show that inference-time computation can improve agent planning, action selection, and error recovery.
However, most existing methods focus on the {policy side} of TTS, assuming resettable or replicable environment sandboxes, but this assumption is difficult to satisfy in shared, persistent, or externally managed environments. Our work addresses this complementary bottleneck by grounding simulation in a single real environment through read-only access, enabling counterfactual trajectory evaluation without speculative writes.

\paragraph{World models for agents.}
World models approximate environment dynamics so that agents can anticipate the consequences of candidate actions. They have been widely studied in model-based reinforcement learning, video prediction, and robotics, where learned models forecast future states or observations for planning and control \citep{dreamv3,genie,dino-wm,v-jepa2}. More recently, LLM-based world models have been used to simulate action outcomes for planning \citep{webdreamer,wma,artis}, or construct executable synthetic environments for agent training \citep{agentwm,simia}. 
Despite these differences, existing approaches generally formulate world modeling as generation under a fixed information set. No formulation allows the simulator to adaptively inspect the current real environment when uncertainty arises during a hypothetical transition. In contrast, we formulate next-observation prediction as an interactive inference process: the simulator identifies missing action-relevant information, acquires it through sequential non-mutating queries, and constructs its prediction from the resulting evidence. This shifts world modeling from passive generation to evidence-driven simulation while preserving the benefits of test-time planning.

\section{Conclusion}

Existing approaches to agentic test-time scaling typically evaluate candidate trajectories through direct execution in resettable environments or independent replicas. This is impractical in realistic shared environments, which are often neither easily reversible nor cheaply replicable. LLM world models avoid speculative writes, but conventional models predict from a fixed, compressed view that may omit transition-relevant state. We therefore introduced \emph{Agentic World Modeling}, which reframes next-observation prediction as multi-turn evidence acquisition. For each hypothetical action, a context-isolated simulator queries the real environment through read-only interfaces and uses executable computation to construct a grounded observation without mutating shared state. This enables parallel counterfactual rollouts over a single non-resettable environment. Across heterogeneous benchmarks, our approach improves simulation fidelity and downstream trajectory selection over single-shot and trained world models, with larger gains as test-time computation increases. These results establish interactive, read-grounded simulation as a practical foundation for safe agentic test-time scaling.

% \section*{Acknowledgement}
% Wendi Li and Sharon Li are supported in part by the AFOSR Young Investigator Program under award number FA9550-23-1-0184, the Office of Naval Research (ONR) under award number N000142612508, the National Science Foundation under awards IIS-2237037 and IIS-2331669, Schmidt Sciences Foundation, Open Philanthropy (now Coefficient Giving), Alfred P. Sloan Fellowship, and gifts from Google and Amazon. Any opinions, findings, conclusions, or recommendations expressed in this material are those of the authors and do not necessarily reflect the views of the National Science Foundation. 

\bibliography{iclr2027_conference}
\bibliographystyle{iclr2027_conference}

\newpage
\appendix
\section{Theoretical Analysis of Adaptive Evidence Acquisition}
\label{app:theory}

This section provides a formal justification for
Proposition~\ref{prop:adaptive-gap}. We show that the adaptive evidence
acquisition enabled by agentic world modeling (AWM) can be exponentially more
information-efficient than non-adaptive context construction under the same
environment-access model.

\paragraph{Adaptive and non-adaptive evidence acquisition.}
Recall that, for predicting a hypothetical transition
$o_{t+1}\sim\mathcal{P}(\cdot\mid h_t,a_t)$, AWM constructs an evidence
trajectory
\[
\xi_i
=
(\tilde a_1,\tilde o_1,\ldots,
 \tilde a_{i-1},\tilde o_{i-1}),
\]
and chooses each read action according to
\[
\tilde a_i
\sim
\pi_\theta(\cdot\mid h_t,a_t,\xi_i).
\]
Thus, the result of one environment read can determine which read should be
performed next.

We compare this with a general class of \emph{non-adaptive context
construction} methods. Given $(h_t,a_t)$, such a method chooses all records to
inspect before observing any of their returned contents. After retrieval, it
may arbitrarily filter, concatenate, or summarize the resulting information
before predicting $o_{t+1}$. This class includes heuristic retrieval,
retrieve-then-summarize methods, and summarization over a predetermined subset
of environment state. Importantly, the selection may depend on $a_t$; the
restriction is only that it cannot depend on intermediate retrieved evidence.
For a fair comparison, both adaptive and non-adaptive methods are charged for
the number of environment reads they perform.

\paragraph{A multi-stage lookup environment.}
We construct a family of transitions in which the information needed for the
next observation is revealed through successive dependencies. Fix an action
$a_t$ that identifies an initial record. Reading this record reveals one
relevant next record among $K$ possible choices. Reading that record again
reveals one among another $K$ possibilities, and this process continues for
$D$ stages. Hence, before any records are inspected, there are
$K^D$ possible final locations.

Let $U=(U_1,\ldots,U_D),
U_i\overset{\mathrm{i.i.d.}}{\sim}\mathrm{Unif}(\{1,\ldots,K\})$ denote the realized sequence of choices. The corresponding final location is therefore uniformly distributed among the $K^D$ possibilities. At this final location, the environment stores $W^\star\sim\mathrm{Unif}(\{0,1\}^{\beta})$, independently of $U$. All other final locations contain information independent of $W^\star$. For the lower-bound construction, we take the next observation to be $o_{t+1}=W^\star$; equivalently, it may be any injective function of $W^\star$. Consequently, faithfully predicting the next observation requires recovering the $\beta$ bits contained in $W^\star$.

\paragraph{Adaptive acquisition.}
The adaptive simulator begins from the initial record specified by $a_t$. Its first returned observation reveals $U_1$, so it queries the corresponding next record. That observation reveals $U_2$, and the procedure repeats until the final record is reached. Thus, the evidence trajectory follows
\[
U_1 \rightarrow U_2 \rightarrow \cdots \rightarrow U_D
\rightarrow W^\star.
\]
The simulator reaches the target after at most $D+1$ reads. Once the final
record has been observed, $W^\star$ is completely determined by the acquired
evidence $\xi^{\mathrm{AD}}$, and therefore
\[
H(W^\star\mid\xi^{\mathrm{AD}})=0.
\]
Equivalently,
\[
I(W^\star;\xi^{\mathrm{AD}})
=
H(W^\star)
=
\beta.
\]
Hence the next observation can be recovered exactly $P_{\mathrm{err}}^{\mathrm{AD}}=0.$ In particular, for any target error $\delta\ge0$,
\begin{equation}
\label{eq:adaptive-budget}
M_{\mathrm{AD}}(\delta)\le D+1.
\end{equation}

\paragraph{Non-adaptive acquisition.}
Now consider any non-adaptive method using $M$ reads. Because all read
locations must be selected before observing their contents, the method cannot
use $U_1,\ldots,U_D$ to follow the realized information path.

Let $E$ denote the event that its predetermined reads include the final record
containing $W^\star$. Since this record is uniformly distributed among
$K^D$ possible final locations, and at most $M$ locations can be inspected,
\begin{equation}
\label{eq:hit-prob}
\Pr(E)
\le
\frac{M}{K^D}.
\end{equation}
This inequality also holds for randomized non-adaptive policies by averaging
over their randomly selected query sets.
If $E$ does not occur, the retrieved evidence contains no information about
$W^\star$ by construction. If $E$ occurs, it can reveal at most all
$H(W^\star)=\beta$ bits. Because the location of $W^\star$ is independent of
its value, we therefore have
\begin{align}
I(W^\star;\xi^{\mathrm{NA}})
\le
\Pr(E)\,H(W^\star) 
\le
\frac{M}{K^D}\beta.
\label{eq:na-mutual-information}
\end{align}
Thus, under the same read budget, a non-adaptive method recovers only a
vanishing fraction of the transition-relevant information as the number of
possible dependency paths grows.

\paragraph{From information to prediction error.}
We next translate the mutual-information bound into a bound on
next-observation prediction accuracy. Since $W^\star$ is uniform over
$2^\beta$ possible values, Fano's inequality gives, for any estimator
$\widehat W$ based on $\xi^{\mathrm{NA}}$,
\[
H(W^\star\mid\xi^{\mathrm{NA}})
\le
1+
P_{\mathrm{err}}^{\mathrm{NA}}\beta,
\]
where
$P_{\mathrm{err}}^{\mathrm{NA}}
=\Pr[\widehat W\neq W^\star]$.
Using
\[
H(W^\star)
=
I(W^\star;\xi^{\mathrm{NA}})
+
H(W^\star\mid\xi^{\mathrm{NA}})
=
\beta
\]
and Eq.~\eqref{eq:na-mutual-information}, we obtain
\begin{align}
P_{\mathrm{err}}^{\mathrm{NA}}
&\ge
1-
\frac{I(W^\star;\xi^{\mathrm{NA}})}{\beta}
-
\frac{1}{\beta}
\ge
1-\frac{M}{K^D}-\frac{1}{\beta}.
\label{eq:na-error}
\end{align}

Therefore, achieving prediction error at most $\delta$ requires
\[
1-\frac{M}{K^D}-\frac{1}{\beta}
\le \delta,
\]
or equivalently
\begin{equation}
\label{eq:na-budget}
M_{\mathrm{NA}}(\delta)
\ge
\left(
1-\delta-\frac{1}{\beta}
\right)K^D.
\end{equation}
Combining Eqs.~\eqref{eq:adaptive-budget}
and~\eqref{eq:na-budget} yields Proposition~\ref{prop:adaptive-gap}:
\[
M_{\mathrm{AD}}(\delta)\le D+1,
\qquad
M_{\mathrm{NA}}(\delta)
\ge
\left(
1-\delta-\frac{1}{\beta}
\right)K^D.
\]
Hence, adaptive evidence acquisition can reduce the required number of
environment accesses from exponential in the dependency depth,
$\Omega(K^D)$, to linear, $O(D)$.

\paragraph{Implications for agentic world modeling.}
The construction isolates the fundamental benefit of AWM: it does not need to determine the complete relevant context \emph{before} interacting with the environment. In large environments, the location of transition-relevant information may itself depend on information discovered during retrieval. A customer record may reveal an account identifier; that account may identify an applicable policy; and the policy may point to an exception needed to determine the resulting observation. A fixed summarizer or heuristic retriever must anticipate these dependencies and distribute its limited access budget across many possible information paths. AWM instead closes the feedback loop $\tilde o_i \longrightarrow \tilde a_{i+1},$ using each retrieved observation to decide what evidence is needed next.

The proposition is a worst-case separation rather than a claim that every environment exhibits an exponential gap. When all transition-relevant information can be identified directly from $(h_t,a_t)$, non-adaptive retrieval can be sufficient. The advantage grows precisely when environment state is large, interdependent, and difficult to anticipate---the regime that motivates our formulation. In such settings, AWM converts a fixed context budget from a requirement to \emph{predict relevance in advance} into a sequential process that can \emph{discover relevance from evidence}, allowing substantially more transition-relevant information to be recovered with the same environment-access budget.

\section{Detailed Experimental Settings}
\label{app:experimental_settings}

This section provides additional details about the benchmarks, baselines, model configurations, test-time scaling protocol, and evaluation metrics used in our experiments.

\subsection{Benchmarks}
\label{app:benchmarks}

We evaluate on three benchmarks that represent complementary forms of agent--environment interaction.

\textbf{$\tau$-bench}~\citep{taubench} evaluates agents in realistic customer-service interactions. We use its \emph{airline} and \emph{retail} domains, containing 50 and 114 tasks, respectively. In each task, the agent converses with an LLM-simulated user while accessing a structured database through domain-specific API tools. The agent must infer the user's intent, comply with the relevant policy document, and perform the required operations. Tasks frequently involve stateful actions, such as modifying a reservation, canceling an order, or issuing a refund. Their outcomes depend jointly on the current database state, applicable policy constraints, and preceding interaction history, making them suitable for evaluating grounded transition prediction.

\textbf{AppWorld}~\citep{appworld} provides a simulated digital ecosystem containing nine applications, 457 APIs, 750 tasks, and approximately 100 fictitious users. We evaluate on 585 tasks divided into Level~1, Level~2, and Level~3 subsets, containing 129, 198, and 258 tasks, respectively. Tasks require the agent to generate and execute code that invokes application APIs, inspect returned values, and coordinate operations across applications. Higher-level tasks generally involve longer dependency chains, more applications, and a greater number of intermediate state changes. Accurate simulation may therefore require retrieving information from several interdependent records rather than relying on a single state description.

\textbf{OfficeBench}~\citep{officebench} evaluates automation in an integrated office environment containing applications such as a word processor, spreadsheet editor, and email client. We use all 300 tasks, grouped by the number of applications involved: 93 single-application tasks, 95 two-application tasks, and 112 three-application tasks. Multi-application tasks require the agent to transfer information across application boundaries and maintain consistency over relatively long interaction horizons. Because actions may modify documents, spreadsheets, or messages, directly executing multiple speculative trajectories in the same workspace can produce persistent side effects and cross-trajectory interference.

\subsection{Compared Methods}
\label{app:baselines}

We compare AWM with two representative world-modeling paradigms. All methods are placed within the same trajectory-sampling and selection framework whenever their interfaces permit, so that the principal difference is how each method predicts the consequence of a hypothetical action.

\textbf{SIMIA}~\citep{simia} represents prompting-based, in-context world modeling. It constructs a textual representation of the environment using heuristically selected information and asks the simulator to generate the next observation in a single inference step. SIMIA cannot acquire additional evidence after observing the prediction query; its prediction is therefore conditioned entirely on the statically constructed prompt.

\textbf{WM-Training}~\citep{qwenworld} represents training-based world modeling. It learns environment dynamics through the three-stage recipe of \citet{qwenworld}, consisting of continued pretraining, supervised fine-tuning, and reinforcement learning. We use the released Qwen-AgentWorld-35B-A3B model as the trained world model. This comparison tests whether adaptive grounding at inference time can complement or outperform environment knowledge acquired through dedicated training.

\subsection{Test-Time Scaling Protocol}
\label{app:tts_protocol}

For each task, the task agent samples \(N\) candidate trajectories using observations predicted by the corresponding world-modeling method. None of the hypothetical mutating actions in these candidate trajectories are applied to the real environment during simulation. After the rollouts terminate, their predicted outcomes are scored by the simulator and the highest-scoring candidate is selected. Only this selected trajectory is then committed to the real environment, and its actual outcome is evaluated using the benchmark's native success criterion. Therefore, all reported task success rates measure real-environment task completion after Best-of-\(N\) selection; the world model's predicted score is used only to select the candidate trajectory.

We set the task-agent sampling temperature to \(1.0\) for all methods and benchmarks with top-$p$ set to $1.0$. For AWM, each observation-prediction subagent is allowed at most six read-only interaction steps per hypothetical transition. The subagent may stop earlier once it determines that the gathered evidence is sufficient. This fixed upper bound controls simulator-side computation while preserving the ability to adapt the number and content of queries to each transition. 
The experiments on Qwen-series models are run with the vLLM server on NVIDIA H200. For other experiments, we directly call the API through Azure Foundry.

\subsection{Environment-access control}
All world-modeling methods are given access to the same underlying application information when constructing their predictions. In particular, the context constructor of baselines can access the same environment records that are exposed to AWM through read-only interfaces; AWM does not receive additional privileged state information. The difference lies in when evidence is selected: the baselines select and serialize their evidence into a fixed context before next-observation prediction, whereas AWM can condition each subsequent read on evidence returned by earlier reads. Our non-adaptive AWM ablation further isolates this distinction by using the same read budget, Python capability, and approximately matched token budget while removing sequential adaptivity.

\section{Supplementary Results}

\begin{table}[t!]
\centering
\small
\setlength{\tabcolsep}{7pt}
\renewcommand{\arraystretch}{1.12}
\caption{Python tool ablation. Success rate (\%) for AWM with and without the Python tool at each rollout budget $N$. Bold indicates the higher score within each model--benchmark pair.}
\label{tab:python-tool-ablation}
\begin{tabular}{@{}llrrrrr@{}}
\toprule
Benchmark & Python tool & \multicolumn{5}{c}{Number of rollouts ($N$)} \\
\cmidrule(lr){3-7}
 & & 1 & 2 & 4 & 8 & 16 \\

\midrule
\multicolumn{7}{l}{\textbf{GPT-5.4}} \\
\addlinespace[2pt]

$\tau$-bench
& With
& \textbf{52.44}
& \textbf{65.85}
& \textbf{71.95}
& \textbf{75.61}
& \textbf{78.66} \\
& Without
& 45.73
& 56.10
& 62.20
& 68.29
& 71.95 \\

\addlinespace[3pt]
AppWorld
& With
& \textbf{52.31}
& \textbf{55.90}
& \textbf{59.49}
& \textbf{62.39}
& \textbf{65.30} \\
& Without
& 45.13
& 50.09
& 54.36
& 58.97
& 60.17 \\

\addlinespace[3pt]
OfficeBench
& With
& \textbf{44.33}
& \textbf{48.67}
& \textbf{50.00}
& \textbf{51.33}
& \textbf{52.67} \\
& Without
& 35.00
& 37.67
& 41.00
& 45.00
& 47.67 \\

\midrule
\multicolumn{7}{l}{\textbf{Qwen3.5-35B-A3B}} \\
\addlinespace[2pt]

$\tau$-bench
& With
& \textbf{64.02}
& \textbf{68.90}
& \textbf{73.78}
& \textbf{76.22}
& \textbf{78.66} \\
& Without
& 60.37
& 64.63
& 69.51
& 71.95
& 73.78 \\

\addlinespace[3pt]
AppWorld
& With
& \textbf{24.62}
& \textbf{28.21}
& \textbf{31.97}
& \textbf{39.32}
& \textbf{45.30} \\
& Without
& 18.97
& 22.05
& 24.96
& 30.94
& 36.07 \\

\addlinespace[3pt]
OfficeBench
& With
& \textbf{46.67}
& \textbf{51.67}
& \textbf{56.67}
& \textbf{59.33}
& \textbf{61.67} \\
& Without
& 41.67
& 46.00
& 51.00
& 54.00
& 56.67 \\

\bottomrule
\end{tabular}
\end{table}
\subsection{Ablation of Program Tools}

The ablation results indicate that the Python tool contributes consistently to AWM's performance across both models and all three benchmarks. Removing it reduces success rates at every rollout budget, with average drops of 8.05, 5.33, and 8.13 percentage points for GPT-5.4 on $\tau$-bench, AppWorld, and OfficeBench, respectively, and corresponding drops of 4.27, 7.28, and 5.33 percentage points for Qwen3.5-35B-A3B. 
The gaps persist at $N=16$, suggesting that increasing the rollout budget alone does not compensate for removing the Python tool. Together, these comparisons support the importance of programmatic execution for effective agent simulation.

\begin{table}[t!]
\centering
\small
\setlength{\tabcolsep}{7pt}
\renewcommand{\arraystretch}{1.10}
\caption{
Scaling performance. Success rate (\%) of AWM at simulator aggregation
levels $M\in\{1,2,4\}$ and rollout budgets $N\in\{1,2,4,8,16\}$.
Benchmark-level rates pool all evaluation tasks across subsets
($164$ for $\tau$-bench, $585$ for AppWorld, and $300$ for OfficeBench),
so each reported percentage corresponds to an integer number of successful
tasks. Real-environment best-of-$N$ (BoN) is the reference ceiling.
Bold indicates the best AWM score for each model, benchmark, and rollout
budget; the reference is excluded from bolding.
}
\label{tab:scaling-performance}
\begin{tabular}{@{}llrrrrr@{}}
\toprule
Benchmark & Method & \multicolumn{5}{c}{Number of rollouts ($N$)} \\
\cmidrule(lr){3-7}
 & & 1 & 2 & 4 & 8 & 16 \\

\midrule
\multicolumn{7}{l}{\textbf{GPT-5.4}} \\
\addlinespace[2pt]

$\tau$-bench
& AWM ($M=1$)
& 52.44 & 65.85 & 71.95 & 75.61 & 78.66 \\
& AWM ($M=2$)
& 56.10 & 70.12 & 78.05 & 81.10 & 84.15 \\
& AWM ($M=4$)
& \textbf{60.37}
& \textbf{74.39}
& \textbf{84.15}
& \textbf{85.98}
& \textbf{89.02} \\
& Real-env BoN
& 64.02 & 78.66 & 90.24 & 91.46 & 94.51 \\

\addlinespace[3pt]
AppWorld
& AWM ($M=1$)
& 52.31 & 55.90 & 59.49 & 62.39 & 65.30 \\
& AWM ($M=2$)
& 55.04 & 60.00 & 64.96 & 68.89 & 71.45 \\
& AWM ($M=4$)
& \textbf{57.61}
& \textbf{63.93}
& \textbf{70.26}
& \textbf{75.21}
& \textbf{77.78} \\
& Real-env BoN
& 71.97 & 79.15 & 85.47 & 90.94 & 93.68 \\

\addlinespace[3pt]
OfficeBench
& AWM ($M=1$)
& 44.33 & 48.67 & 50.00 & 51.33 & 52.67 \\
& AWM ($M=2$)
& 45.33 & 50.33 & 52.67 & 54.33 & 56.00 \\
& AWM ($M=4$)
& \textbf{46.67}
& \textbf{52.33}
& \textbf{55.33}
& \textbf{57.67}
& \textbf{59.00} \\
& Real-env BoN
& 47.67 & 54.00 & 58.00 & 60.67 & 62.33 \\

\midrule
\multicolumn{7}{l}{\textbf{Qwen3.5-35B-A3B}} \\
\addlinespace[2pt]

$\tau$-bench
& AWM ($M=1$)
& 64.02 & 68.90 & 73.78 & 76.22 & 78.66 \\
& AWM ($M=2$)
& 65.24 & 69.51 & 75.00 & 78.05 & 79.27 \\
& AWM ($M=4$)
& \textbf{66.46}
& \textbf{70.73}
& \textbf{76.83}
& \textbf{79.27}
& \textbf{79.88} \\
& Real-env BoN
& 67.68 & 71.34 & 78.05 & 79.88 & 81.10 \\

\addlinespace[3pt]
AppWorld
& AWM ($M=1$)
& 24.62 & 28.21 & 31.97 & 39.32 & 45.30 \\
& AWM ($M=2$)
& 27.52 & 40.34 & 46.15 & 53.16 & 57.26 \\
& AWM ($M=4$)
& \textbf{30.26}
& \textbf{52.48}
& \textbf{60.51}
& \textbf{66.84}
& \textbf{69.40} \\
& Real-env BoN
& 33.16 & 64.62 & 74.70 & 80.68 & 81.37 \\

\addlinespace[3pt]
OfficeBench
& AWM ($M=1$)
& 46.67 & 51.67 & 56.67 & 59.33 & 61.67 \\
& AWM ($M=2$)
& 48.00 & 52.67 & 59.00 & 62.00 & 64.33 \\
& AWM ($M=4$)
& \textbf{49.33}
& \textbf{54.00}
& \textbf{61.67}
& \textbf{64.33}
& \textbf{66.67} \\
& Real-env BoN
& 52.33 & 58.33 & 64.00 & 67.00 & 69.33 \\

\bottomrule
\end{tabular}
\end{table}

\subsection{Effect of Scaling Simulation}

Table~\ref{tab:scaling-performance} shows consistent improvements from
scaling both the rollout budget $N$ and the simulator aggregation level $M$ across both models and all three benchmarks. At $N=16$, increasing $M$ from 1 to 4 improves GPT-5.4's success rate by 10.37, 12.48, and 6.33 percentage points on $\tau$-bench, AppWorld, and OfficeBench, respectively, reaching 89.02\%, 77.78\%, and 59.00\%. The corresponding gains for Qwen3.5-35B-A3B are 1.22, 24.10, and 5.00 percentage points, reaching 79.88\%, 69.40\%, and 66.67\%.
Across settings, increasing simulator-side aggregation consistently improves downstream task success and moves performance closer to the real-environment BoN reference, although the remaining gap varies substantially across model--benchmark combinations.

\subsection{Direct Evaluation of Next-Observation Prediction Accuracy}

We directly evaluate world-model fidelity using trajectories generated by Qwen in the real environment. At each step, the policy samples an action $a_t$ from the current trajectory history $h_t$. We then execute $a_t$ in the real environment to obtain the ground-truth observation $o_{t+1}$, while independently asking the world model to predict $\hat{o}_{t+1}$ from the same $(h_t,a_t)$. Prediction accuracy is measured by comparing $\hat{o}_{t+1}$ with the corresponding real observation $o_{t+1}$. To prevent simulation errors from compounding across steps, we append the \emph{real} observation $o_{t+1}$ to the trajectory history before sampling the next action. This protocol therefore isolates one-step next-observation prediction quality while evaluating the model on actions encountered along realistic agent trajectories. On Qwen3.5-35B-A3B, we randomly sample 64 tasks from each benchmark, approximately balanced across its subsets, and evaluate every next-observation prediction encountered along the resulting real-environment trajectories. AWM achieves next-observation prediction accuracies of 41.1\% on AppWorld and 82.2\% on OfficeBench, compared with 6.87\% and 5.94\% for SIMIA, respectively. The comparison explicitly shows that our approach can indeed enhance the reliability of next observation prediction.

\begin{table}[t]
\centering
\caption{
Ablation with non-adaptive evidence acquisition. We report task success rate (\%) on each benchmark subset at BoN@16. The non-adaptive variant specifies all read operations before observing their returned evidence, while adaptive AWM conditions subsequent reads on previously acquired evidence. Both variants use the same maximum budget of six reads, Python capability, and approximately matched token budget.
}
\label{tab:nonadaptive_ablation}
\resizebox{\textwidth}{!}{
\begin{tabular}{llcccccccc}
\toprule
& &
\multicolumn{2}{c}{$\tau$-bench} &
\multicolumn{3}{c}{AppWorld} &
\multicolumn{3}{c}{OfficeBench} \\
\cmidrule(lr){3-4}
\cmidrule(lr){5-7}
\cmidrule(lr){8-10}
Model & Evidence acquisition
& Airline & Retail
& L1 & L2  & L3 
& 1App & 2Apps & 3Apps \\
\midrule

\multirow{2}{*}{GPT-5.4}
& Non-adaptive AWM
& 74.00 & 68.42
& 82.17 & 55.05 & 37.98
& 58.06 & 53.68 & 19.64 \\
& \textbf{AWM}
& \textbf{82.00} & \textbf{77.19}
& \textbf{87.60} & \textbf{65.66} & \textbf{53.88}
& \textbf{63.44} & \textbf{63.16} & \textbf{34.82} \\

\midrule

\multirow{2}{*}{Qwen3.5-35B-A3B}
& Non-adaptive AWM
& 68.00 & 71.93
& 58.91 & 37.88 & 17.44
& 66.67 & 65.26 & 26.79 \\
& \textbf{AWM}
& \textbf{76.00} & \textbf{79.82}
& \textbf{64.34} & \textbf{48.48} & \textbf{33.33}
& \textbf{72.04} & \textbf{74.74} & \textbf{41.96} \\

\bottomrule
\end{tabular}
}
\end{table}
\subsection{Ablation with Non-Adaptive Evidence Acquisition}
\label{app:nonadaptive}

To isolate the contribution of adaptive evidence acquisition, we construct a non-adaptive variant of AWM. Given $(h_t,a_t)$, the simulator must specify all read operations before observing any of their returned results. These reads are then executed together, and the complete retrieved evidence is provided to the same simulator for next-observation prediction. We retain the same maximum budget of six read operations, Python execution capability, and approximately matched overall token budget. The primary difference is therefore whether a later read can condition on evidence returned by earlier reads.

Table~\ref{tab:nonadaptive_ablation} shows that removing this adaptivity consistently reduces downstream task success across both model backbones and all evaluation subsets. More importantly, the degradation generally becomes larger as tasks require more compositional environment interaction. On AppWorld, the gap between adaptive and non-adaptive acquisition for GPT-5.4 increases from 5.43 percentage points on Level 1 to 10.61 points on Level 2 and 15.89 points on Level 3. A similar trend appears for Qwen3.5-35B-A3B, with corresponding gaps of 5.43, 10.61, and 15.89 points. OfficeBench exhibits the same pattern: for GPT-5.4, the gap grows from 5.38 points on single-application tasks to 9.47 points on two-application tasks and 15.18 points on three-application tasks, with a closely matching trend for Qwen3.5-35B-A3B.

This behavior is consistent with the motivation behind adaptive evidence acquisition. For relatively simple transitions, the simulator can often anticipate the relevant records directly from $(h_t,a_t)$, making a preselected set of reads reasonably effective. In more complex tasks, however, relevant information is more likely to be distributed across multiple dependent records or applications, so the result of one read can determine what should be inspected next. A non-adaptive simulator must anticipate these dependencies before observing any evidence, whereas AWM can progressively resolve them by conditioning each subsequent query on the evidence acquired so far. The increasing performance gap on harder subsets therefore provides empirical support for the adaptive advantage analyzed in Section~\ref{sec:theory}.

\section{Case Study}
As shown in Figure~\ref{fig:case-study-SIMIA-AWM}, Figure~\ref{fig:case-study-quantities}, and Figure~\ref{fig:case-study-relationships}, the six case studies illustrate how AWM produces more reliable simulated observations than SIMIA across AppWorld and OfficeBench. SIMIA relies on a fixed prompt to represent environment information; when necessary records are omitted, it can incorrectly predict empty directories, zero transactions, missing files, or blank document contents. In contrast, AWM retains or retrieves the relevant evidence and processes it through Python, recovering task-critical information such as transaction amounts, playlist memberships, spreadsheet values, and notification recipients. Although some AWM predictions retain ordering or formatting errors, they preserve the substantive information required for downstream decisions. These examples highlight the reliability advantage of grounding simulation in accessible records and executable computation rather than relying solely on fixed-context prediction.
  
\definecolor{bad}{HTML}{9C3737}
\definecolor{good}{HTML}{17695C}
\definecolor{ink}{HTML}{243344}
\lstdefinestyle{casecode}{basicstyle=\ttfamily\small,breaklines=true,columns=fullflexible,keepspaces=true,showstringspaces=false,aboveskip=4pt,belowskip=4pt}
\lstset{style=casecode}
\tcbset{casepanel/.style={enhanced,boxrule=0.6pt,arc=2mm,left=7pt,right=7pt,top=5pt,bottom=5pt,colback=white,fonttitle=\bfseries\normalsize}}

\begin{figure*}[t]
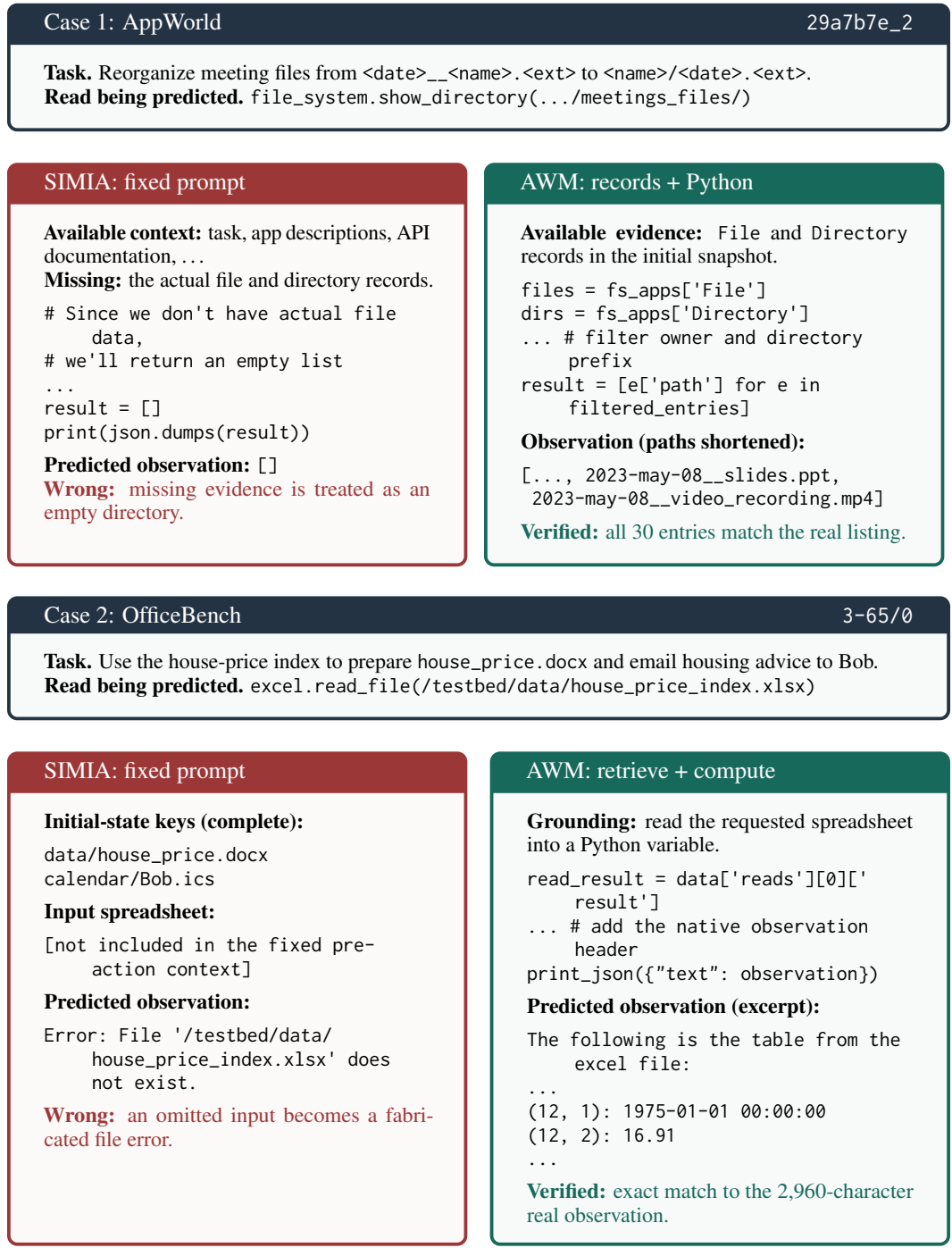

\centering
\begin{tcolorbox}[colframe=ink,colback=ink!3,title={Case 1: AppWorld \hfill \texttt{29a7b7e\_2}}]
\small\textbf{Task.} Reorganize meeting files from
\texttt{<date>\_\_<name>.<ext>} to \texttt{<name>/<date>.<ext>}.\par
\textbf{Read being predicted.} \texttt{file\_system.show\_directory(.../meetings\_files/)}
\end{tcolorbox}
\begin{minipage}[t]{0.488\linewidth}\vspace{0pt}
\begin{tcolorbox}[colframe=bad,colback=bad!3,equal height group=caseone,title={SIMIA: fixed prompt}]
\small\textbf{Available context:} task, app descriptions, API documentation, \ldots\par
\textbf{Missing:} the actual file and directory records.
\begin{lstlisting}
# Since we don't have actual file data,
# we'll return an empty list
...
result = []
print(json.dumps(result))
\end{lstlisting}
\textbf{Predicted observation:} \texttt{[]}\par
\textcolor{bad}{\textbf{Wrong:} missing evidence is treated as an empty directory.}
\end{tcolorbox}
\end{minipage}\hfill
\begin{minipage}[t]{0.488\linewidth}\vspace{0pt}
\begin{tcolorbox}[colframe=good,colback=good!3,equal height group=caseone,title={AWM: records + Python}]
\small\textbf{Available evidence:} \texttt{File} and \texttt{Directory} records in the initial snapshot.
\begin{lstlisting}
files = fs_apps['File']
dirs = fs_apps['Directory']
... # filter owner and directory prefix
result = [e['path'] for e in filtered_entries]
\end{lstlisting}
\textbf{Observation (paths shortened):}
\begin{lstlisting}
[..., 2023-may-08__slides.ppt,
 2023-may-08__video_recording.mp4]
\end{lstlisting}
\textcolor{good}{\textbf{Verified:} all 30 entries match the real listing.}
\end{tcolorbox}
\end{minipage}
\vspace{7pt}

\begin{tcolorbox}[colframe=ink,colback=ink!3,title={Case 2: OfficeBench \hfill \texttt{3-65/0}}]
\small\textbf{Task.} Use the house-price index to prepare \texttt{house\_price.docx}
and email housing advice to Bob.\par
\textbf{Read being predicted.} \texttt{excel.read\_file(/testbed/data/house\_price\_index.xlsx)}
\end{tcolorbox}
\begin{minipage}[t]{0.488\linewidth}\vspace{0pt}
\begin{tcolorbox}[colframe=bad,colback=bad!3,equal height group=casetwo,title={SIMIA: fixed prompt}]
\small\textbf{Initial-state keys (complete):}
\begin{lstlisting}
data/house_price.docx
calendar/Bob.ics
\end{lstlisting}
\textbf{Input spreadsheet:}
\begin{lstlisting}
[not included in the fixed pre-action context]
\end{lstlisting}
\textbf{Predicted observation:}
\begin{lstlisting}
Error: File '/testbed/data/house_price_index.xlsx' does not exist.
\end{lstlisting}
\textcolor{bad}{\textbf{Wrong:} an omitted input becomes a fabricated file error.}
\end{tcolorbox}
\end{minipage}\hfill
\begin{minipage}[t]{0.488\linewidth}\vspace{0pt}
\begin{tcolorbox}[colframe=good,colback=good!3,equal height group=casetwo,title={AWM: retrieve + compute}]
\small\textbf{Grounding:} read the requested spreadsheet into a Python variable.
\begin{lstlisting}
read_result = data['reads'][0]['result']
... # add the native observation header
print_json({"text": observation})
\end{lstlisting}
\textbf{Predicted observation (excerpt):}
\begin{lstlisting}
The following is the table from the excel file:
...
(12, 1): 1975-01-01 00:00:00
(12, 2): 16.91
...
\end{lstlisting}
\textcolor{good}{\textbf{Verified:} exact match to the 2,960-character real observation.}
\end{tcolorbox}
\end{minipage}
\caption{\textbf{Fixed-context omissions undermine simulation reliability.}
We show two predictions on AppWorld and Officebench comparing a SIMIA prediction with a AWM one. Ellipses, shortened paths, and reflowed lines improve readability.}
\label{fig:case-study-SIMIA-AWM}
\end{figure*}
\begin{figure*}[t]
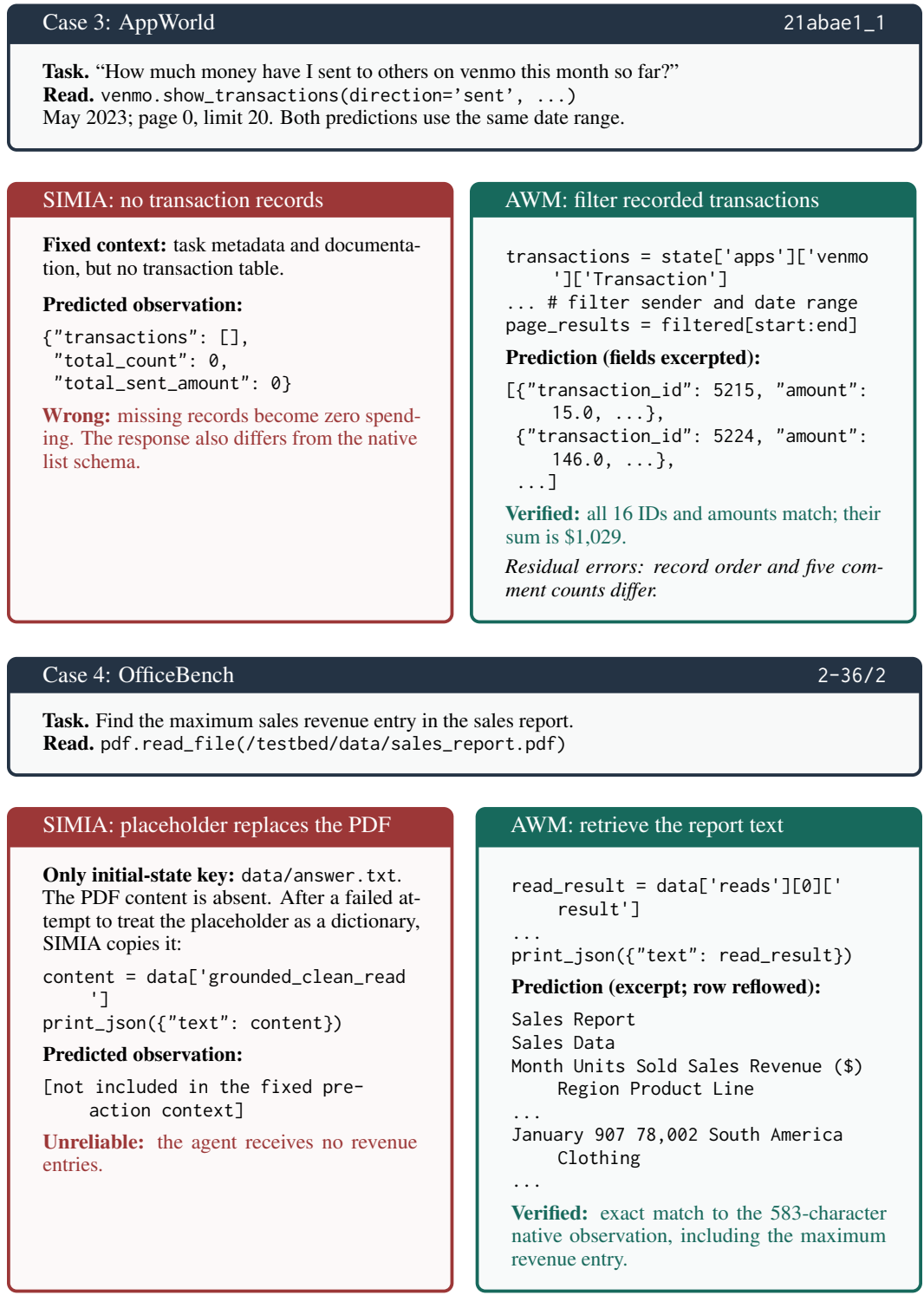

\centering
\begin{tcolorbox}[colframe=ink,colback=ink!3,title={Case 3: AppWorld \hfill \texttt{21abae1\_1}}]
\small\textbf{Task.} ``How much money have I sent to others on venmo this month so far?''\par
\textbf{Read.} \texttt{venmo.show\_transactions(direction='sent', ...)}\par
May 2023; page 0, limit 20. Both predictions use the same date range.
\end{tcolorbox}
\begin{minipage}[t]{0.488\linewidth}\vspace{0pt}
\begin{tcolorbox}[colframe=bad,colback=bad!3,equal height group=casethree,title={SIMIA: no transaction records}]
\small\textbf{Fixed context:} task metadata and documentation, but no transaction table.
\par\medskip\textbf{Predicted observation:}
\begin{lstlisting}
{"transactions": [],
 "total_count": 0,
 "total_sent_amount": 0}
\end{lstlisting}
\textcolor{bad}{\textbf{Wrong:} missing records become zero spending. The response also differs from the native list schema.}
\end{tcolorbox}
\end{minipage}\hfill
\begin{minipage}[t]{0.488\linewidth}\vspace{0pt}
\begin{tcolorbox}[colframe=good,colback=good!3,equal height group=casethree,title={AWM: filter recorded transactions}]
\begin{lstlisting}
transactions = state['apps']['venmo']['Transaction']
... # filter sender and date range
page_results = filtered[start:end]
\end{lstlisting}
\small\textbf{Prediction (fields excerpted):}
\begin{lstlisting}
[{"transaction_id": 5215, "amount": 15.0, ...},
 {"transaction_id": 5224, "amount": 146.0, ...},
 ...]
\end{lstlisting}
\textcolor{good}{\textbf{Verified:} all 16 IDs and amounts match; their sum is \$1,029.}\par
\smallskip\textit{Residual errors: record order and five comment counts differ.}
\end{tcolorbox}
\end{minipage}
\vspace{9pt}
\begin{tcolorbox}[colframe=ink,colback=ink!3,title={Case 4: OfficeBench \hfill \texttt{2-36/2}}]
\small\textbf{Task.} Find the maximum sales revenue entry in the sales report.\par
\textbf{Read.} \texttt{pdf.read\_file(/testbed/data/sales\_report.pdf)}
\end{tcolorbox}
\begin{minipage}[t]{0.488\linewidth}\vspace{0pt}
\begin{tcolorbox}[colframe=bad,colback=bad!3,equal height group=casefour,title={SIMIA: placeholder replaces the PDF}]
\small\textbf{Only initial-state key:} \texttt{data/answer.txt}.\par
The PDF content is absent. After a failed attempt to treat the placeholder as a dictionary, SIMIA copies it:
\begin{lstlisting}
content = data['grounded_clean_read']
print_json({"text": content})
\end{lstlisting}
\textbf{Predicted observation:}
\begin{lstlisting}
[not included in the fixed pre-action context]
\end{lstlisting}
\textcolor{bad}{\textbf{Unreliable:} the agent receives no revenue entries.}
\end{tcolorbox}
\end{minipage}\hfill
\begin{minipage}[t]{0.488\linewidth}\vspace{0pt}
\begin{tcolorbox}[colframe=good,colback=good!3,equal height group=casefour,title={AWM: retrieve the report text}]
\begin{lstlisting}
read_result = data['reads'][0]['result']
...
print_json({"text": read_result})
\end{lstlisting}
\small\textbf{Prediction (excerpt; row reflowed):}
\begin{lstlisting}
Sales Report
Sales Data
Month Units Sold Sales Revenue ($) Region Product Line
...
January 907 78,002 South America Clothing
...
\end{lstlisting}
\textcolor{good}{\textbf{Verified:} exact match to the 583-character native observation, including the maximum revenue entry.}
\end{tcolorbox}
\end{minipage}
\caption{\textbf{Recovering the quantities needed for the task.} SIMIA predictions are compared with AWM computation for AppWorld and OfficeBench. Native reads independently verify the displayed facts. Case 3 establishes correct transaction IDs and amounts, not an exact full response; the residual errors are shown explicitly. Case 4 is an exact response match. Excerpts omit unrelated fields and use editorial comments after ellipses.}
\label{fig:case-study-quantities}
\end{figure*}
\begin{figure*}[t]
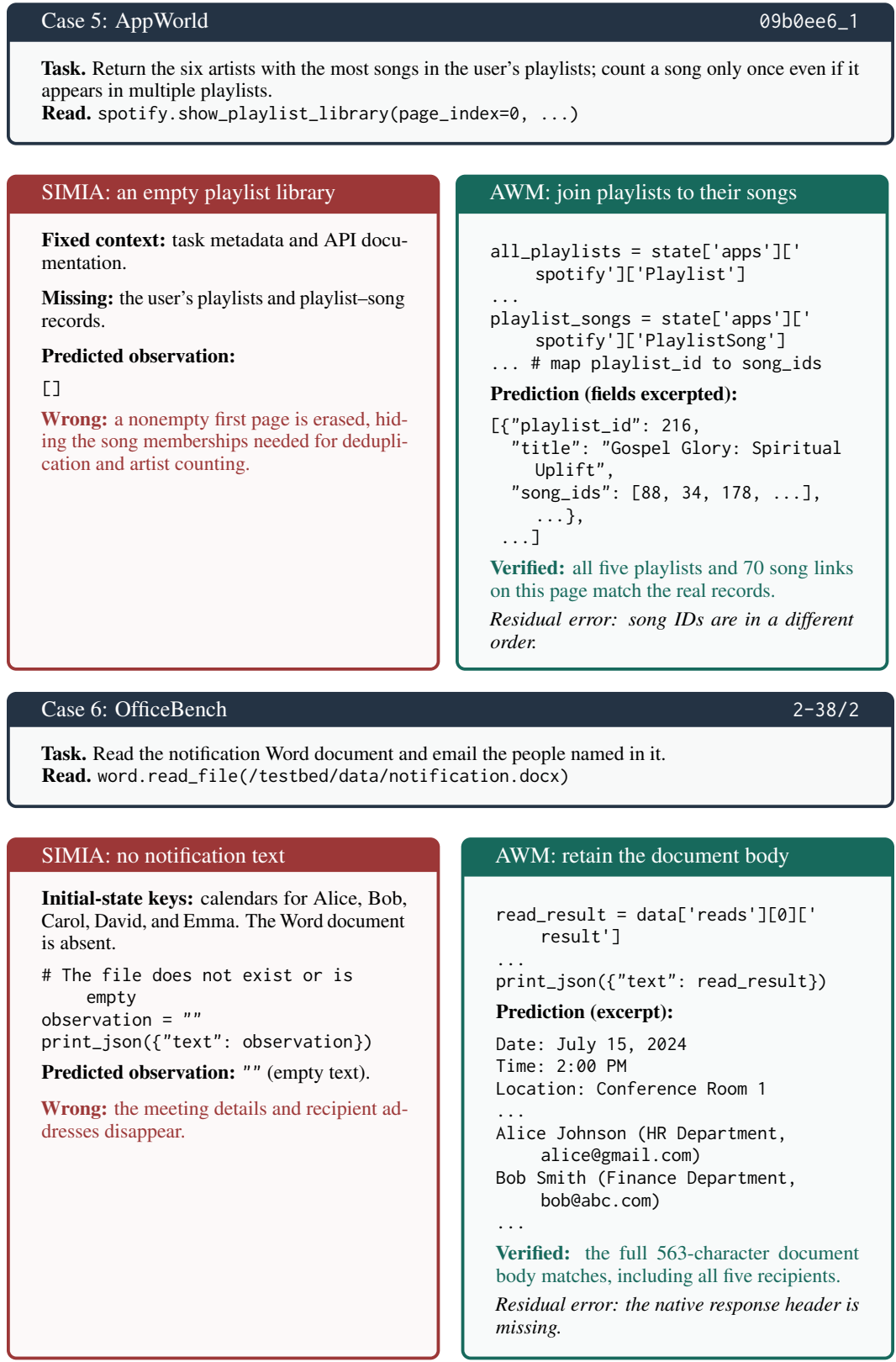

\centering
\begin{tcolorbox}[colframe=ink,colback=ink!3,title={Case 5: AppWorld \hfill \texttt{09b0ee6\_1}}]
\small\textbf{Task.} Return the six artists with the most songs in the user's playlists;
count a song only once even if it appears in multiple playlists.\par
\textbf{Read.} \texttt{spotify.show\_playlist\_library(page\_index=0, ...)}
\end{tcolorbox}
\begin{minipage}[t]{0.488\linewidth}\vspace{0pt}
\begin{tcolorbox}[colframe=bad,colback=bad!3,equal height group=casefive,title={SIMIA: an empty playlist library}]
\small\textbf{Fixed context:} task metadata and API documentation.\par
\medskip\textbf{Missing:} the user's playlists and playlist--song records.\par
\medskip\textbf{Predicted observation:}
\begin{lstlisting}
[]
\end{lstlisting}
\textcolor{bad}{\textbf{Wrong:} a nonempty first page is erased, hiding the song memberships needed for deduplication and artist counting.}
\end{tcolorbox}
\end{minipage}\hfill
\begin{minipage}[t]{0.488\linewidth}\vspace{0pt}
\begin{tcolorbox}[colframe=good,colback=good!3,equal height group=casefive,title={AWM: join playlists to their songs}]
\begin{lstlisting}
all_playlists = state['apps']['spotify']['Playlist']
...
playlist_songs = state['apps']['spotify']['PlaylistSong']
... # map playlist_id to song_ids
\end{lstlisting}
\small\textbf{Prediction (fields excerpted):}
\begin{lstlisting}
[{"playlist_id": 216,
  "title": "Gospel Glory: Spiritual Uplift",
  "song_ids": [88, 34, 178, ...], ...},
 ...]
\end{lstlisting}
\textcolor{good}{\textbf{Verified:} all five playlists and 70 song links on this page match the real records.}\par
\smallskip\textit{Residual error: song IDs are in a different order.}
\end{tcolorbox}
\end{minipage}
\vspace{4pt}
\begin{tcolorbox}[colframe=ink,colback=ink!3,title={Case 6: OfficeBench \hfill \texttt{2-38/2}}]
\small\textbf{Task.} Read the notification Word document and email the people named in it.\par
\textbf{Read.} \texttt{word.read\_file(/testbed/data/notification.docx)}
\end{tcolorbox}
\begin{minipage}[t]{0.488\linewidth}\vspace{0pt}
\begin{tcolorbox}[colframe=bad,colback=bad!3,equal height group=casesix,title={SIMIA: no notification text}]
\small\textbf{Initial-state keys:} calendars for Alice, Bob, Carol, David, and Emma. The Word document is absent.
\begin{lstlisting}
# The file does not exist or is empty
observation = ""
print_json({"text": observation})
\end{lstlisting}
\textbf{Predicted observation:} \texttt{""} (empty text).\par
\medskip\textcolor{bad}{\textbf{Wrong:} the meeting details and recipient addresses disappear.}
\end{tcolorbox}
\end{minipage}\hfill
\begin{minipage}[t]{0.488\linewidth}\vspace{0pt}
\begin{tcolorbox}[colframe=good,colback=good!3,equal height group=casesix,title={AWM: retain the document body}]
\begin{lstlisting}
read_result = data['reads'][0]['result']
...
print_json({"text": read_result})
\end{lstlisting}
\small\textbf{Prediction (excerpt):}
\begin{lstlisting}
Date: July 15, 2024
Time: 2:00 PM
Location: Conference Room 1
...
Alice Johnson (HR Department, alice@gmail.com)
Bob Smith (Finance Department, bob@abc.com)
...
\end{lstlisting}
\textcolor{good}{\textbf{Verified:} the full 563-character document body matches, including all five recipients.}\par
\smallskip\textit{Residual error: the native response header is missing.}
\end{tcolorbox}
\end{minipage}
\caption{\textbf{Retaining relationships and document contents.}  SIMIA predictions are compared with AWM for AppWorld and OfficeBench. Ellipses omit content for readability. Together, these examples show why putting simulator information in a fixed prompt can omit necessary evidence.}
\label{fig:case-study-relationships}
\end{figure*}

\end{document}